\documentclass[runningheads]{llncs}
\usepackage{graphicx}
\usepackage[utf8]{inputenc}
\usepackage{lineno,hyperref}
\usepackage{amssymb}
\usepackage{graphicx}
\usepackage{dcolumn}
\usepackage{bm}
\usepackage{epstopdf}
\usepackage{amsmath}
\usepackage{textcomp} 
\usepackage{multirow}

\usepackage{times}  
\usepackage{helvet} 
\usepackage{courier}  
\usepackage[misc]{ifsym}
\usepackage{hyperref}
\newcommand{\comment}[1]{}
\begin{document}
\title{Long-Term Pipeline Failure Prediction Using Nonparametric Survival Analysis}
%
%
%

\author{Dilusha Weeraddana\inst{1}(\Letter)\and
Sudaraka MallawaArachchi\inst{2}\and
Tharindu Warnakula\inst{2} \and
Zhidong Li\inst{3}\and
Yang Wang \inst{3}}


\authorrunning{D. Weeraddana et al.}

\institute{\textsuperscript{1}Data61-The Commonwealth Scientific and Industrial Research Organisation (CSIRO),
  Australia\\
  \textsuperscript{2}Monash University,
  Australia\\
  \textsuperscript{3}University of Technology Sydney,
  Australia}

%
%
\maketitle              
\begin{abstract}

Australian water infrastructure is more than a hundred years old, thus has begun to show its
age through water main failures. Our work concerns  approximately half a million pipelines across major Australian cities that deliver water to houses and businesses, serving over five million customers. Failures on these buried assets cause damage to properties and water supply disruptions. We applied Machine Learning techniques to find a cost-effective solution to the pipe failure problem in these Australian cities, where on average 1500 of water main failures occur each year. To achieve this objective, we construct a detailed picture and understanding of the behaviour of the water pipe network by  developing a Machine Learning model to assess and predict the failure likelihood of water main breaking using historical failure records, descriptors of pipes and other environmental factors. Our results indicate that our system incorporating a nonparametric survival analysis technique called \lq Random Survival Forest\rq \ outperforms several popular algorithms and expert heuristics in long-term prediction. In addition, we construct a statistical inference technique to quantify the uncertainty associated with the long-term predictions.

\end{abstract}
\section{Introduction}\label{sec:introduction}
\begin{figure*}
  \includegraphics[width=\textwidth]{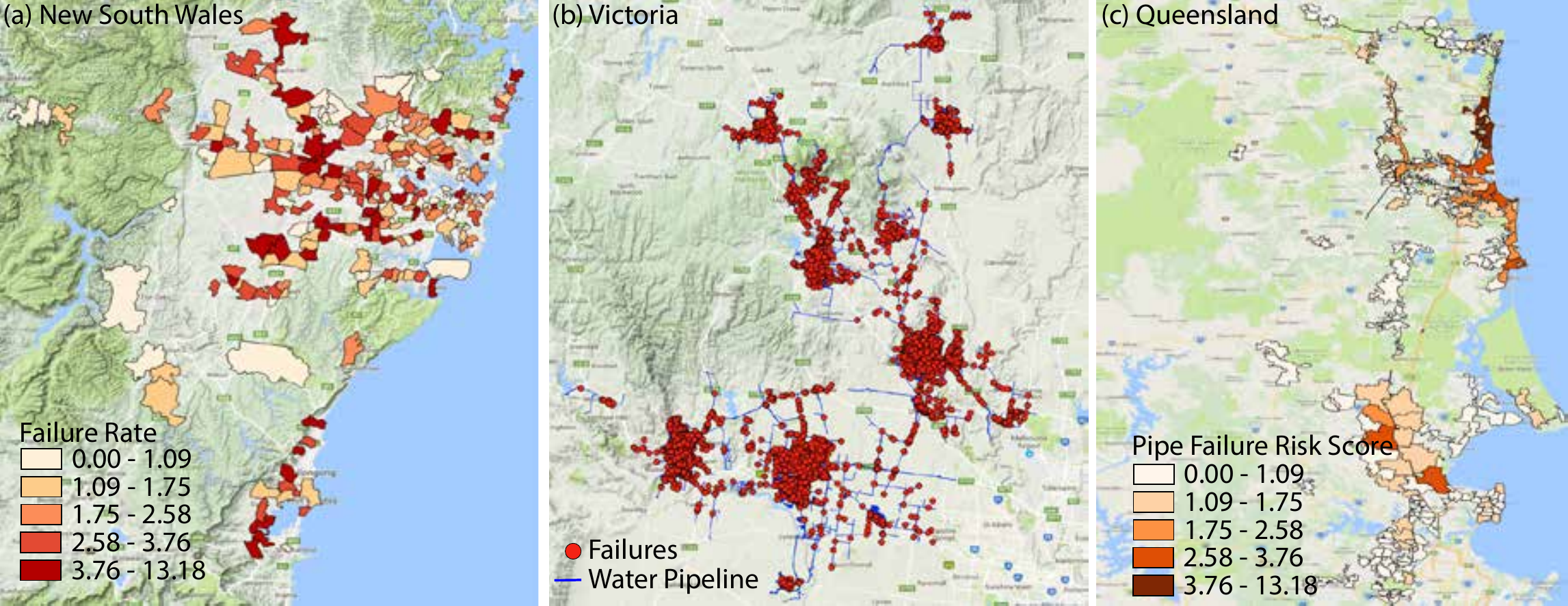}
  \caption{\label{fig:title}Water pipeline failure statistics in three major regions of Australia, using three metrics; (a) pipeline failure rate (number of failures per year per 100 km), (b) failure count on the water pipeline network, (c) risk factor (failure likelihood $\times$ consequence of the failure) across the network.}

\end{figure*}

The degradation of urban water mains causes a major problem in urban engineering in Australia. The most common measures of pipeline breakage are the frequency of the water pipe breaks (breaks per 100 km per year) and the criticality factor of the breakage. Pipeline failure rate varies widely, as it depends on various factors, such as pipe material, pipe diameter and various other environmental and operational conditions. The maintenance and renewal of water mains demand high financial investments. Moreover, direct inspection of all water mains in a distribution system is extremely expensive. Therefore, a cost-effective break mitigation technique such as a prediction model that allows one to predict the water mains failure, would reduce the negative customer impact and the cost to serve.  Consequently, this proactive maintenance model elaborates an optimized strategy for water mains maintenance and rehabilitation.

\subsection{The water pipeline failure problem}
\begin{figure}[h]
	\centering
	\includegraphics[width=\textwidth]{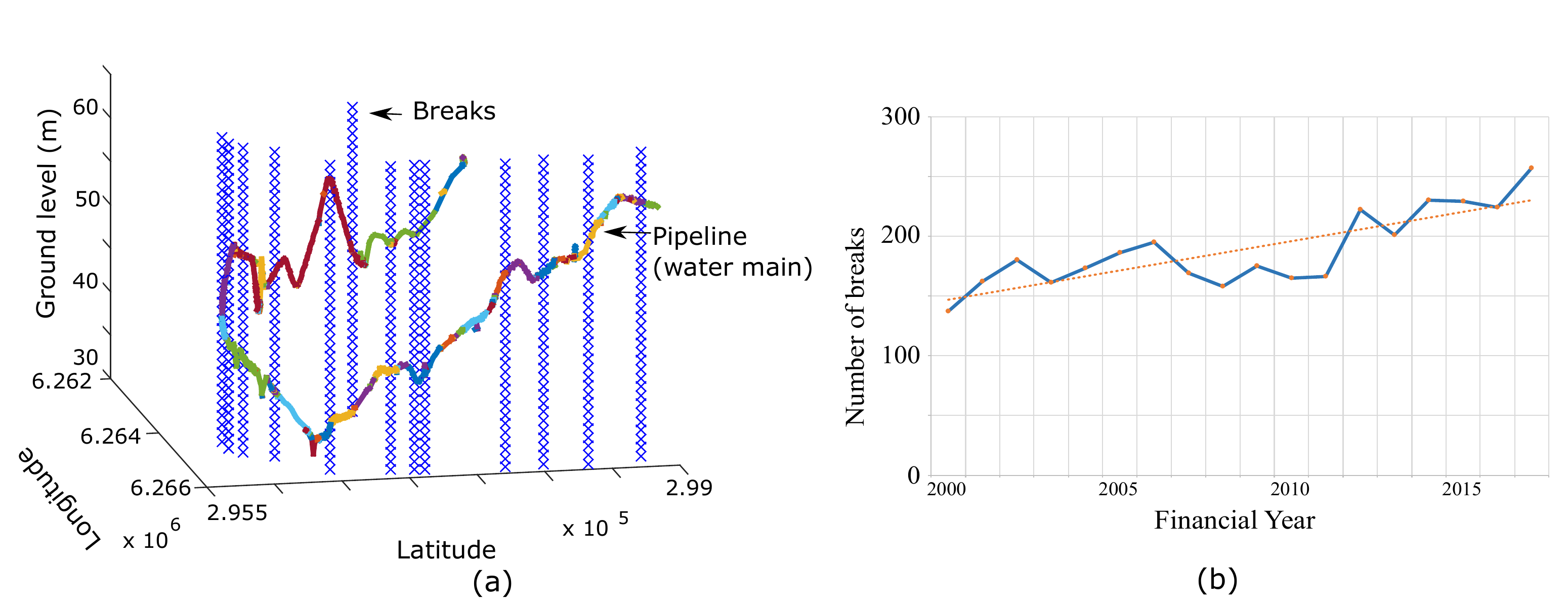}
	\caption{\label{gl_breaks} (a) 3D schematic of a water pipeline which comprises of various individual pipes. X axis, Y axis, and Z axis denote latitude, longitude and the ground level of the water main, respectively. Vertical blue lines represent the breaks occurred in this water main since 2000, (b) Increment in number of critical pipe breaks over the recent years in Sydney, Australia.}
\end{figure}

This study concerns the failure analysis of the water pipeline network of three major cities located in three different states in Australia, namely: New South Wales (NSW), Victoria (VIC), and Queensland (QLD). The water network includes a total of 500,000 pipelines. The oldest pipes were laid in 1890 in Romsey, VIC and surrounding suburbs. The total length of this pipeline network is over 30,000 km.  As depicted in Figure \ref{gl_breaks} (a), a water main comprises of several pipes and each pipe comprises several pipe nodes buried in various ground levels.

Water pipe failures are mainly studied using three different metrics, namely: failure rate, failure count and the risk factor associated with the failure. Figure \ref{fig:title}(a) shows the pipeline failure rates in major cities of NSW, Australia. The failure rate is the number of asset failures per 100 km per year. Higher failure rates are illustrated in darker red spots, and it clearly shows higher failure rates are not localized to one area, they are spread across the state. Breakages in the water main network in the region west of VIC is shown in Figure \ref{fig:title}(b). Figure \ref{fig:title}(c) illustrates the risk distribution of pipeline network in south-east QLD.  Across the entire region under our study, an average number of 1500 pipe failures occur each year, causing water supply disruptions and myriads of property and environmental damages. Figure \ref{gl_breaks} (b) shows the increasing trend in breakage of critical pipes (each water utility has its own method to identify the criticality of a water pipe depending on the risk associated with its breakage) in NSW from 2000 to 2017.

%

\subsection{Related work}
There has been a lot of work in recent years on pipe failure prediction in  water infrastructure, ranging from physical models \cite{CRONIN2002,gould2011} to machine learning models \cite{li2014,Bang2018,Simon2017}, \cite{Kumar2018} and the combination of both \cite{shi2015}.

Machine learning based pipe failure forecasting dates back to 1979 \cite{shamir1979}, where authors developed a forecasting technique to study how the number of breaks would change with time if the pipes were not replaced. In that study, authors used a Poisson model based on the age of the pipes. However, research carried out afterwards shows that the age is not the only factor that causes the pipeline failure. In fact, some of the very old pipes function more robust than their newly laid counterparts. Furthermore, the prediction of water main breaks has been studied widely using statistical based approaches, such as Poisson regression and Weibull models \cite{asnaashari2009,vanrenterghem2003}. Most recently, tree-based Machine Learning techniques have been used to analyse water pipe breakages in Syracuse, USA \cite{Kumar2018}, and QLD, Australia \cite{Bin2017}. The former study shows that Gradient Boosting (GB) outperforms other methods when predicting high risk city blocks. The work reported in \cite{Dilu2019} uses a combination of Random Forest (RF) and linear regression to predict the long-term pipe failure likelihood for water and sewer pipes in QLD, Australia.

Although numerous research have been conducted on forecasting water pipeline failures, open questions still exist regarding the intricate relationship among the major factors causing pipe failure and their long-term effects on the life-time of a pipe. This may vary depending on the environment (weather, soil, ground level, pressure, etc.) that the pipe is laid in and the pipe maintenance approach of each water utility. Thus, prediction of the water main breaks becomes a complicated task due to their low failure rate and high cost of inspection, which have led to a sparse historical data.

Most of the research found in the literature predict short-term failure forecasting, which spans 1-3 years into the future \cite{Kumar2018}. However, water utilities require long-term estimations for the structural deterioration of water mains to effectively plan the renewal of water distribution systems and to develop a risk based investment decisions for capital interventions \cite{Dilu2020}. Therefore, the main objective of this work is to investigate pipe failure factors and develop a long-term prediction model.

\subsection{Our Contribution}
There are approximately 22850 km, 5900 km, and 2000 km of water mains in the city of Sydney, South-east QLD and the region West of Melbourne, respectively. These pipeline networks comprise approximately 500,000 water pipes serving major residential and industrial cities in Australia. We implement a machine learning based prediction model using the Random Survival Forest (RSF) to identify future pipe failure likelihoods for water main asset in these Australian cities.  Firstly, we generate failure likelihood of each pipe using RSF, as it is fully nonparametric and does not impose a restrictive structure on data distribution or how the variables should be combined \cite{Wey2015}. If the relationship between the independent variables and the dependent variable is complex with non linear interactions, then the RSF algorithm is capable of capturing these intricate relationships \cite{ishwaran2008,Nasejje2017}. In our model, the predictions were validated by separating the data into training and testing samples. Afterwards, a derived list is generated and evaluated on the testing data. We further compare the results from RSF with a variety of other approaches, such as GB, RF and Weibull. Water authorities are often interested in obtaining a confidence interval for the predictions we produce. This is due to the fact that, pipe failure  predictions suffer from various sources of error, such as the variations in weather conditions, new infrastructure developments, root clogs caused by near by trees, and many other sources, which are caused by the inherent stochastic and nonlinear characteristics of water pipe failures. In order to quantify the uncertainty in failure forecasting effectively, we have generated the uncertainty interval for the long-term prediction by treating RSF as quantile regression forests. As a result, for each point that is predicted with a RSF, we provide the perceived uncertainty of that prediction.

In the past, RSF has been employed in various medical related research \cite{dietrich2016random,miao2015}. However, to the best of our knowledge, this is the first model applied on pipe failure problem embracing the quantile regression forests \cite{meinshausen2006quantile} for uncertainty estimation, and proven on real-world datasets collected from multiple water authorities.

Our data analytical model provides a projection of the likelihood of future pipe failures. These likelihoods, along with the consequence of failures, are currently being used in  current investment planning of each of these Australian water utilities, to make risk based investment decisions for capital interventions. Thus, our contributions help the water asset renewal programs  to reduce the catastrophic consequence of water main failures and the cost to customers.

\subsection{Preliminary}
\subsubsection{Survival and hazard functions}
The survival function, $S(t)$ is a non-increasing function, which provides the probability that a subject will survive past time $t$ \cite{cox2018,klein2006}.
\begin{eqnarray}{\label{SF}}
S(t) = Pr(T > t)=\int_{t}^{\infty} f(u) du\nonumber\\
\end{eqnarray}
Here, $T$ is a continuous random variable with the probability density function: $f(u)$, or more generally, $T$ represents the waiting time until the occurrence of an event. In our scenario, the survival function illustrates the probability that a particular pipe survives past a given time.
The hazard function describes the event rate,
\begin{eqnarray}{\label{SF}}
\lambda(t)=\lim_{\delta t\to 0}\frac{Pr(t<T \leq t+\delta t|T>t)}{\delta t}\nonumber\\
\end{eqnarray}
\begin{eqnarray}{\label{SF_2}}
S(t)=\exp^{-\lambda(t)}\nonumber\\
\end{eqnarray}

The Cumulative Hazard Function (CHF) provides the accumulated risk up to time $t$,
\begin{eqnarray}{\label{SF}}
\mu(t)=\int_{0}^{t} \lambda(u) du\nonumber\\
\end{eqnarray}
$\mu(t)$ can be seen as the sum of the risks accumulating from duration $0$ to $t$. Thus, these functions are of intrinsically pivotal in forecasting about the condition of a pipe which has survived a certain time period.

\subsubsection{Random Survival Forest}
An extension of RF to the domain of survival analysis enhances its value greatly. In survival analysis, many different regression modeling strategies, such as Cox regression and Poisson regression, can be applied to predict the survival likelihoods. Extending the RF approach \cite{Breiman2001} to survival analysis provides an alternative way to build a robust asset failure prediction model. This technique safely omits the need to impose parametric or semi-parametric constraints on the underlying distributions and allows for an accurate prediction \cite{ishwaran2008,miao2015}.

RSF consists of arbitrarily grown survival trees. Using independent bootstrap samples, each tree is grown by randomly selecting a subset of variables at each node and then splitting the node using the candidate variable that maximizes survival difference between daughter nodes. The tree is grown until saturation is reached due to the condition of each terminal node having no fewer than $d_{0} > 0$ unique deaths (in our case, this referred to the number of pipe breakages). The output of each tree may be estimated as the CHF for each case, the estimator for which is the Nelson–Aalen estimator for the terminal node in which the case ends up \cite{ishwaran2008},
\begin{eqnarray}{\label{SF}}
\mu(t)=\sum_{t_{j}\leq t}\frac{d_{j}}{Y_{j}},\nonumber\\
\end{eqnarray}
where $t_{j}$ are the ordered pipe failure times for the terminal node. $d_{j}$ and $Y_{j}$ are the number of pipe failures and pipes at risk (number of pipes in the terminal node that are functioning) at time $t_{j}$ in the terminal nodes. However, in our model, instead of the CHF, we derive an estimate of the survival probability for each terminal node using the Kaplan-Meier estimator \cite{kaplan1958nonparametric} given by,

\begin{eqnarray}{\label{SF}}
S(t)=\prod_{t_{j}\leq t}\left(1-\frac{d_{j}}{Y_{j}}\right).\nonumber\\
\end{eqnarray}

Given the CHF or survival estimate from a tree, an ensemble average is performed over the entire forest to produce the final prediction.

\section{Data analytic model for pipeline failure prediction}
\subsection{Data extraction and pre-processing}
There are three main data sources used as the inputs to the analytical model, :
\begin{itemize}
	\item Network data: describes water main information such as asset number, installation date, material, size.
	\item Work order data: describes water main failure information such as asset number, failure date, location, and failure type (burst, fitting, leak).
	\item External data: includes information in addition to assets, such as weather data from the Bureau of Meteorology and census data from the Australian Bureau of Statistics, soil data, pressure data, pipe ground level data, etc.
\end{itemize}
The above data should be sufficiently accurate for the intended use, so a data quality review has been undertaken based on three key characteristics:

\begin{itemize}
	\item  Completeness: this is a statistical analysis that does not allow empty values.
	\item  Validity: this is a statistical analysis that removes invalid values.
	\item  Consistency: make sure that the data obtained from all the sources are consistent with each other.
\end{itemize}

The quality review demonstrates that the data is sufficient and accurate for further analysis. Accordingly, this process allows to establish a comprehensive data file with complete information for each asset that can be used as an important input for further analysis. Moreover, when information is gathered from multiple sources, prior to the adoption of advanced analytic techniques, it is essential to match the failure records with the network data and identify gaps in the datasets. In addition, environmental and demographic factors need to be matched with the network data. Specifically, failure records and information are assigned to the corresponding assets based on the work order number.

\subsection{Factor analysis}
Once the data is pre-processed, the next step is to identify the factors that cause pipeline breakage and compare their relative impact on the network based on the water network information. Factor analysis measures the correlation between asset performance based on the comprehensive data and a large range of factors (including environmental, demographic and asset specific factors) \cite{rajeev2014,gould2011}. While a significant amount of literature exists on the pipeline failure causes, this step is critical to discerning which of these causes would be the most important for each water utility. The asset performance is based on failure rate, which is the number of asset failures per 100 km per year. Both single factor analysis and multi-factor analysis have been performed to identify the possible driving factors. The asset performance is not usually related to only one factor, so it is essential to measure the correlation based on multiple factors.

\begin{figure}[h]
	\centering
	\includegraphics[width=\textwidth]{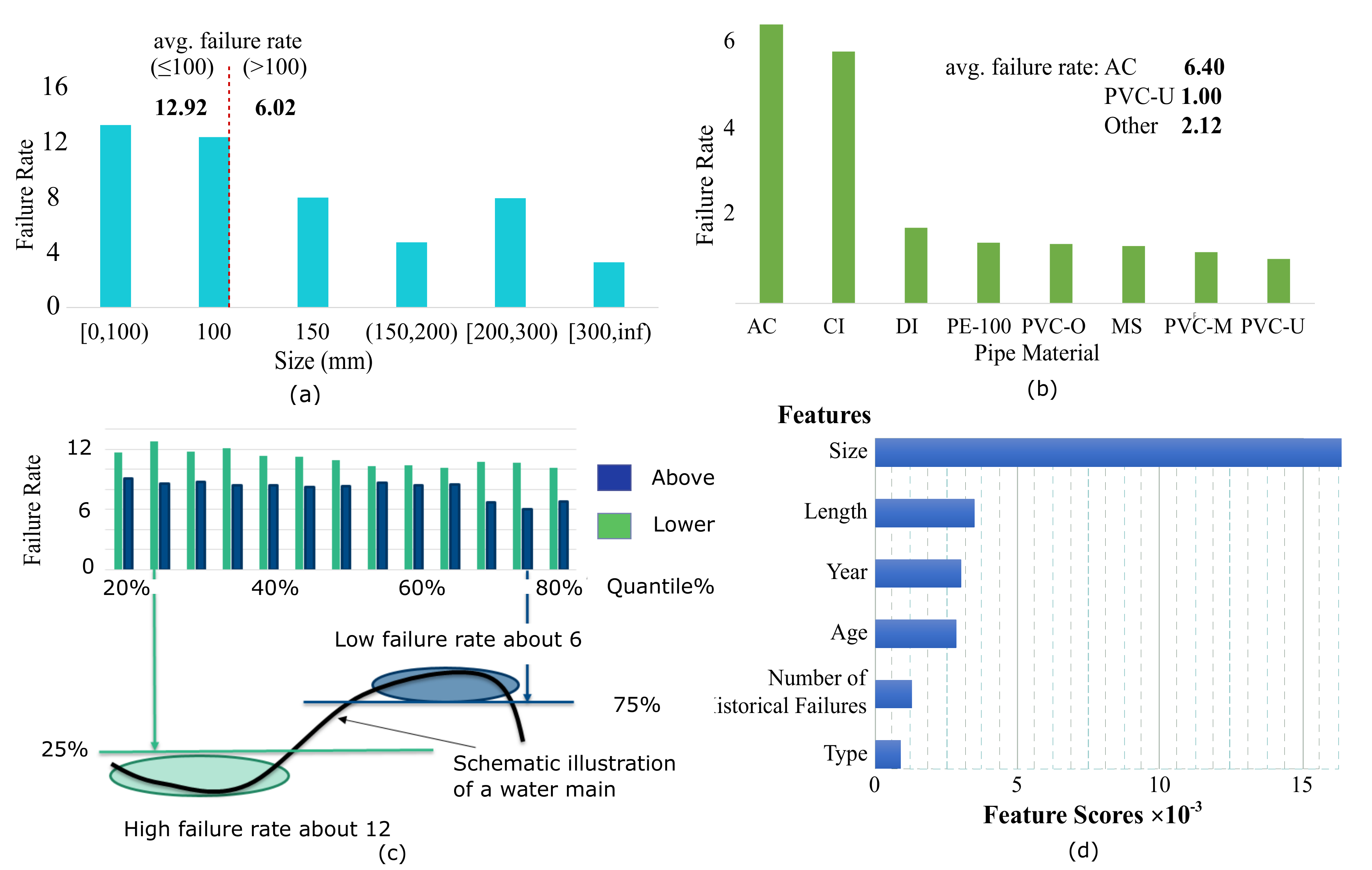}
	\caption{\label{size_type_gl} Factor analysis examples: (a) Failure rate of water mains based on pipe size in pipelines located region west of Melbourne, (b) AC materials are more prone to break in QLD, (c) Factor analysis of the ground level of a water main and how it affects the pipe breakage in pipelines located region west of Sydney. (d) Feature importance scores for the VIC dataset, computed using the mutual entropy gain method.}
\end{figure}

For example, within operational factors, AC water mains were found to break more often than others in the regions of QLD as shown in Figure \ref{size_type_gl} (b). It was also found that water mains  with diameters less than 100 mm exhibit higher failure rates, compared to larger pipes (see Figure \ref{size_type_gl} (a)). Moreover, a quantitative study on the ground level of water main and its impact on the pipe breakage is shown in Figure \ref{size_type_gl} (c). It can be observed that the failure rate of pipes laid in the bottom 25\% of ground level is twice higher than the pipes laid in the top 25\% of ground level (above 75\% of quantile).

To quantify the amount of pipe failure information stored in each of the
features in isolation, we calculate the mutual information between
the \lq Pipe Failure\rq \ parameter and each feature (we have selected a basic set of asset specific features which are common to all three states). The data from all three states display a very similar dependence of
the failures on the predictor variables. Therefore, the resulting
information scores for the VIC dataset are presented in Figure \ref{size_type_gl} (d). Pipe size (or diameter) shares
the highest amount of mutual information with failures while pipe
type has the least effect on failures. In general, all predictors by
themselves display very low levels of mutual information indicating
that by themselves, they do not predict failures sufficiently well.
However, as we shall show later, the six features in unison will
provide us with an excellent prediction model of pipe failures.

To this end, we also identified the potential advantages of analysing the factors causing pipe failures in different datasets across various Australian regions. We have been working with a few water utilities and identifying the differences and the commonalities among these various datasets allow us to improve our knowledge in developing the prediction framework.


\subsection{Long-term failure prediction}
This phase involves predicting future water pipe failure probabilities. We framed this scenario as determining the likelihood of failure on each given pipe. The failure prediction is generated by training the RSF model on historical failure records and other factors, such as pipe material, pipe laid year, pipe diameter, etc. This trained model produces a survival probability score for each water main asset for each years into the future.

The RSF model utilized in this work uses data on the history of water pipeline network across major Australian cities. It specifically uses the failure history of pipes (the response) and their characteristics (the predictor variables). The response variable includes the minimum of the survival time: $T_{i}$, the right censoring time $C_{i}$ and $\Delta_{i} = \Im\{T_{i} \leq C_{i}\}$ which is the censoring value indicating a pipe has failed ($\Delta_{i}=1$) or was right-censored ($\Delta_{i} = 0$). The predictor variables $X_{i}=(X_{i}^{1},...,X_{i}^{N})$ for respective pipe, $i$ consists of both continuous variables, such as age, previous failures, as well as qualitative variables, such as pipe material and pipe size.

\subsubsection{Prediction uncertainty}
We construct a simple, yet effective, statistical inference technique to quantify the uncertainty associated with the predictions generated by supervised learning ensembles. Here, we employ quantile regression forests in survival trees generated by the RSF model. The concept behind the Random survival quantile regression forests is, instead of recording the mean value of response variables in each tree leaf in the forest, record all observed responses in the leaf. The prediction can then be calculated as the mean of the response variables, as well as the full conditional distribution of response values for every $x$. Using the distribution, the prediction intervals for new instances can be generated  by employing the appropriate percentiles of the distribution.

Following \cite{Ishwaran07}, the high-level description of the algorithm used in this work, along with the procedure for determining uncertainty, an be given as follows:
\begin{enumerate}
	\item Ascertain the training year range and the prediction year range of pipe failure observations. A training data file is created on average for eight observation years of pipeline data. Each observation year contains information of all the pipes in the network, with an indication of whether a particular pipe has failed in that observation year or not. The observation year range for the training data is selected and restricted(e.g. 2005-2010).
	\item $N$ number of bootstrap samples are pulled from the training dataset by excluding on average $37\%$ of the data, which is referred to as out-of-bag (OOB) data.
	\item A survival tree is developed for each bootstrap sample. At each node of the tree, a $p$ number of candidate variables are randomly selected. The node is split using the candidate variable that maximizes survival difference between daughter nodes.
	\item  Grow the tree to full size under the constraint that a terminal node should have no less than $d_{0} > 0$ unique deaths.
	\item Using OOB data, the prediction error for the ensemble survival is calculated.	
\item Calculate survival probability for the predicting data range of observation years (e.g. 2011-2025) by recording all observed responses in the leaf, and obtaining conditional probability distribution of the response variable for every given set of predictor variables ($x$) of each pipe. Using the distribution,  create prediction intervals for new instances by using the appropriate percentiles of the distribution to calculate the lower and upper bounds of the prediction uncertainty.
\end{enumerate}

\section{Case study}
We study the pipeline failure data from three major Australian states: VIC, NSW and QLD. Each of these three datasets were generated based on the results of observations made on pipelines in each observation year. As an example, the data statistics for selected laid year groups for VIC are presented in Table 1. This highlights the different dynamics associated in training and testing datasets for each laid year. For VIC and NSW, data spanning observations from 2000 to 2017 were available while for QLD, only data for observations from 2013 to 2017 were available. The key information recorded at these observations is represented as a boolean variable recording whether a failure was detected at the time. We also use auxiliary data regarding each of the observed pipes as input parameters to predict failures into the future. The full list of features used in our modelling is reported in the Figure \ref{size_type_gl}. We use the age of the pipe observed, the year in which it was laid in, the material that the  pipe is made of, the number of previous failures and the size (diameter) of the pipe as predictor variables.

\begin{figure}
	\centering
	\includegraphics[width = \textwidth]{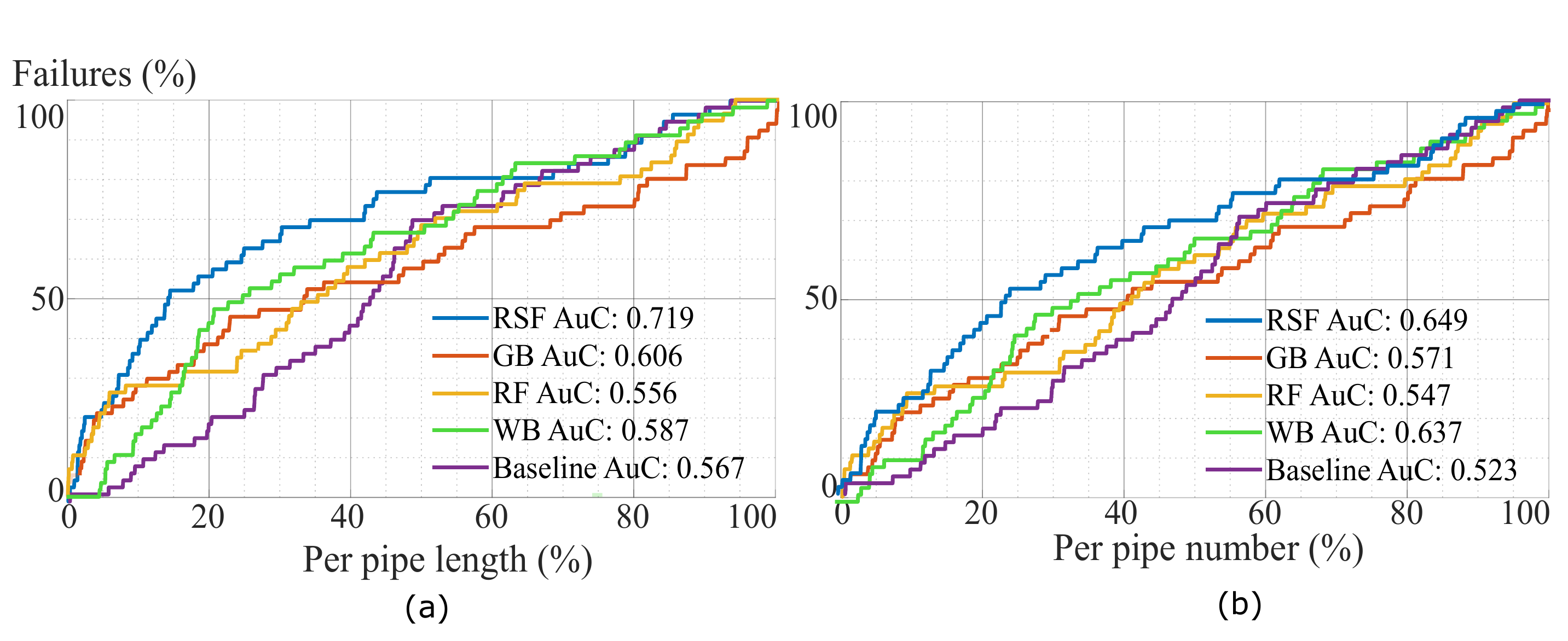}
	\caption{\label{ROCMelb} ROC curves generated based on the predictions made by each technique for the state of VIC for the year 2017. Observation data collected in and before year 2010 was used for the training task. Therefore, the predictions illustrated here are made 7 years in advance. (a) depicts the ROC curves based on the total pipe length and (b) the number of pipes, respectively.}
\end{figure}

\begin{figure}
	\centering
	\includegraphics[width =\textwidth]{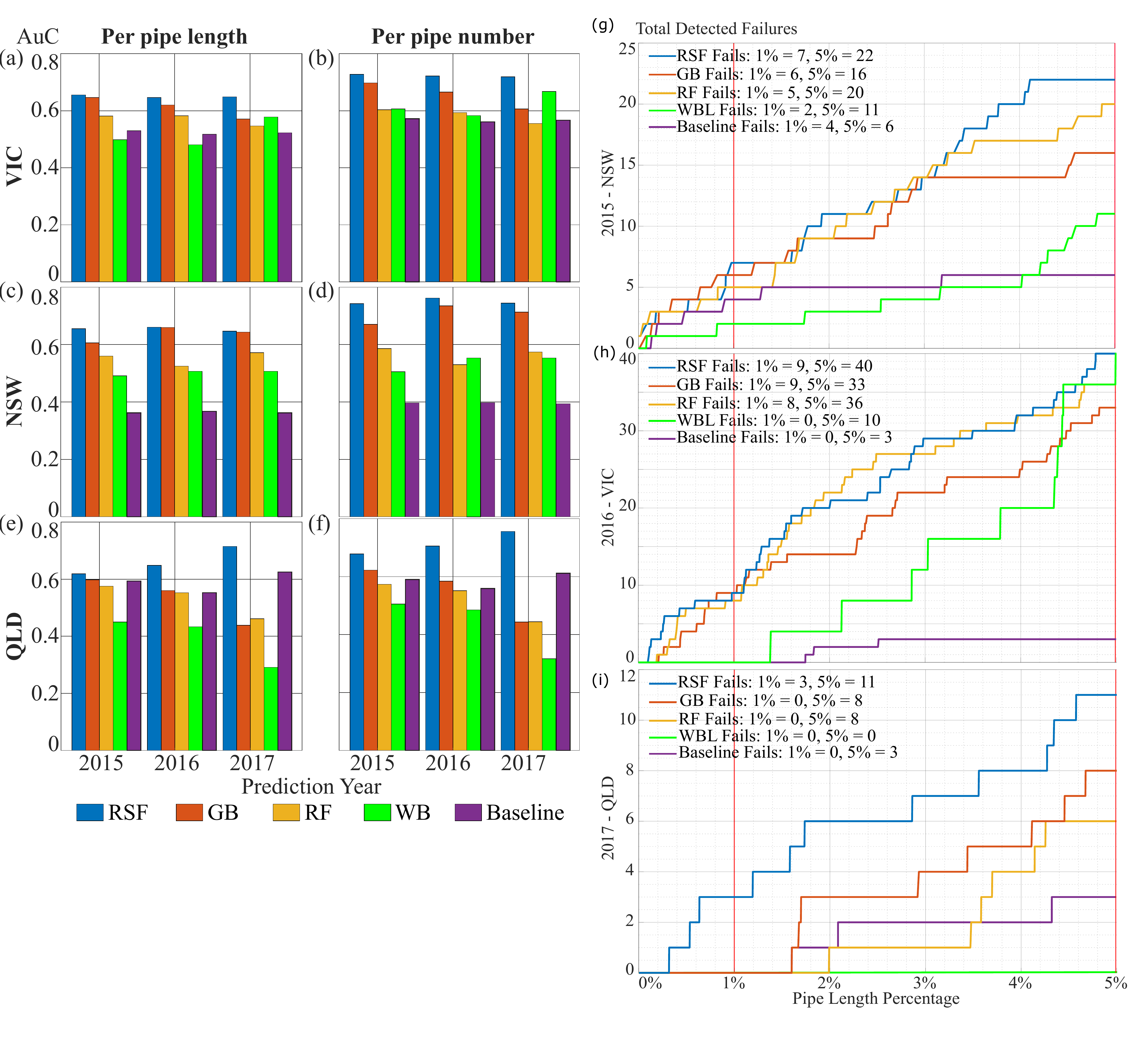}
	\caption{\label{AuCFig} Complete collection of bar plots depicting the AUC values generated for all prediction scenarios. (a) and (b) bar plots show the AUC values obtained for the VIC dataset for the (a) length based and (b) pipe number based ROC curves, respectively. Similarly, (c) and (d) plots correspond to the NSW dataset and (e) and (f) plots correspond to the QLD dataset.\\
Inspection of the top 1\% and 5\% of ranked pipe length in (a) NSW - 2015, (h) VIC - 2016 and (i) QLD - 2017. In each case, the failures correctly identified using the predictions made by RSF (blue), GB (orange) and RF (yellow) techniques are shown therein. The two red vertical lines across each plot identifies the inspection points at 1\% and 5\% of total pipe length. The numerical failure count is indicated in the top left hand corner of each figure for convenience.}
\end{figure}

\subsection{Model Setting}
Using this information, we perform a comprehensive comparison of performance between the Random Survival Forest technique and  other widely used machine learning and statistical algorithms along with a baseline predictor. For VIC and NSW, we use the observations from 2000-2010 to train our machine learning models and only the data from the year 2013 for QLD. The probabilities of failure for pipes observed in the years 2015-2017 are then calculated and compared to the actual recorded observation. This provides all our methods a common benchmark to be compared against.

The algorithms we choose to measure RSF against are, GB technique \cite{friedman2002stochastic}, RF regression \cite{Breiman2001} and Weibull model \cite{vanrenterghem2003}. GB is a machine learning technique that iteratively improves modelling using weak predictors \cite{moisen2006}.  In RF, independently drawn random sub-samples of the complete dataset are used to build an ensemble of regression trees. For all three algorithms (RSF, GB, RF), we use 100 trees when training the model. Additionally, we predicted failure rates for each dataset using a 2-parameter Weibull model. We fitted the Weibull model to the \emph{age at first failure} distribution of all pipes in each dataset, and the computed parameters were used to estimate the probability of pipe failure by aging all pipes according to the prediction year. Further to this,  the baseline predictor we use is the number of previous failures of a given pipe. We assign a higher probability of failure to the pipes with a history of a higher number of failures.



\begin{figure}
	\centering
	\includegraphics[width =\textwidth]{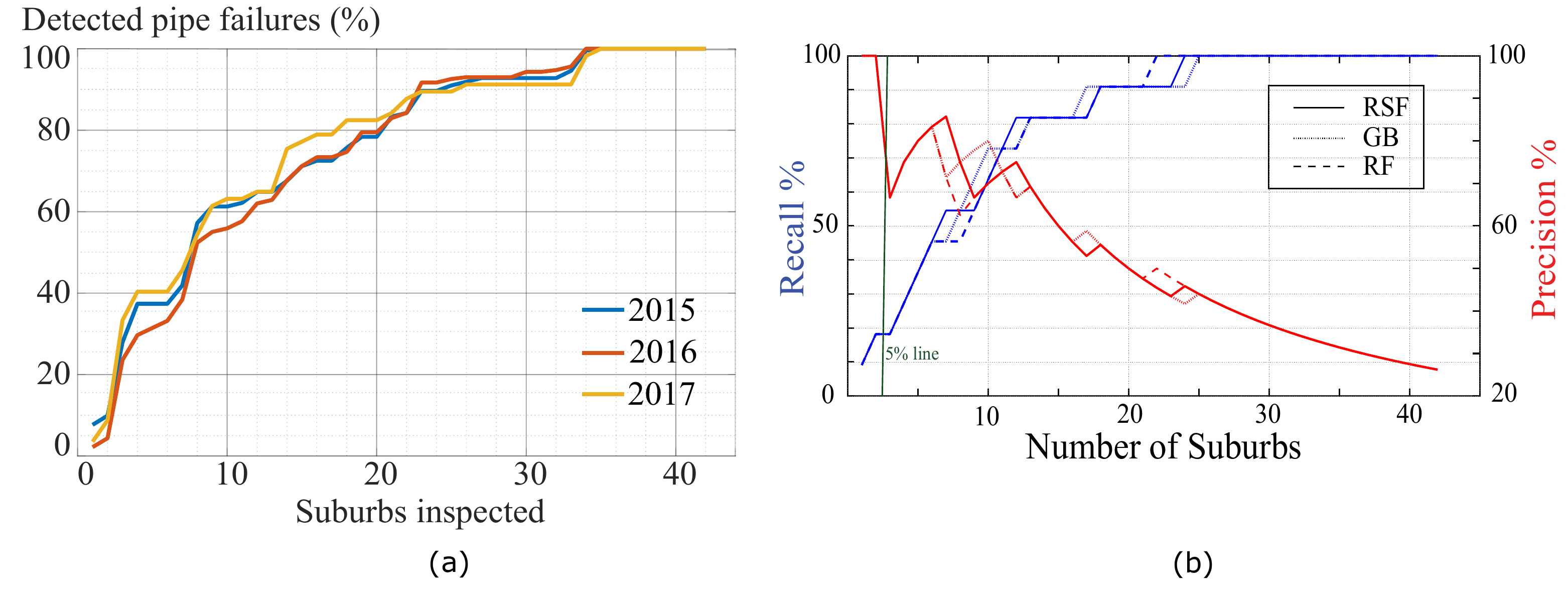}
	\caption{\label{SuburbModel} (a) Suburb level model verification for the VIC pipe dataset. (b) Critical suburb prediction recall (blue) and precision(red) percentages for VIC suburbs in the year 2017. At 5\% (out of a total of 42 suburbs), the recall percentage is at 18.18\% and the precision is at 96.67\% for all three techniques considered. The RSF is plotted as a solid line with the RF as a dashed line and the GB method response as a dotted line. This pattern of similar recall and precision curves across all three techniques and all years is evident especially at lower population levels.}
\end{figure}

To easily compare the predictions, we plot the Receiver Operator Characteristic (ROC) curves of the predictions made by each predictor and compare them through the Area Under the Curve (AUC). Firstly, the prediction model was trained using the pipeline features and the failure data. Then the calibrated model was applied to predict the survival probability for each pipe from year 2015 to 2017. Then the pipes were ranked according to the survival probability of each pipe. Using the ranked list, actual failures from the lowest to the highest probability are accumulated (cumulative sum of failures). The percentage of detected failures is plotted against the percentage of inspected pipe lengths, and the percentage of inspected number of pipes. Predictors that reach higher true positive rates while maintaining low false positive rates are preferred for predicting purposes and correspondingly. We also note that we use two separate methods for generating the ROC. One method defines true positive and false positive rates based on the number of pipes predicted to fail correctly or falsely, while the other defines false positive rate as the total length of pipes incorrectly predicted to fail while retaining the usual definition of true positive rate. We term these pipe based ROC and length based ROC, respectively. While the pipe based ROC is the more natural definition of the two, Water Management authorities consider failures per unit length as an important parameter and the length based ROC accounts for this explicitly.


\subsection{Experimental Results and Discussion}
In order to quantify their prediction performance, we extensively studied the ROC curves generated by each machine learning technique under various scenarios. One such instance is shown in Figure \ref{ROCMelb}, where the complete set of ROC curves generated  by each technique are plotted together. These curves are based on the predictions made for the year 2017 for the VIC pipe dataset. Pipe observations made in and before the year 2010 were used for training purposes and the predictions are made for a time period that is 7 years into the future (year 2017). The length based ROC curve for RSF clearly demonstrates a $\approx5\%$ prediction enhancement over the other two techniques, whereas the pipe number based ROC curve demonstrates a  $\approx10\%$ prediction enhancement.

The bar plots in Figure \ref{AuCFig} further demonstrate the superior prediction capabilities of the RSF technique compared to GB, RF and Weibull techniques. For the VIC pipe dataset, all predictions made using the RSF technique show better prediction results when compared with the GB and RF and Weibull.
RSF  outperforms both RF and GB by a considerable margin in the 2017 prediction. A similar observation can be made for both NSW and QLD pipe datasets; predictions made for a year further away from the last year in the training dataset (2010 for VIC and NSW, 2013 for QLD) tend to show better accuracy when predicted using the RSF technique. Furthermore, it is noted that there are some rare instances where the GB and RF techniques marginally outperform the predictions made by the RSF technique. This behavior is particularly observed for the NSW pipe dataset. This is because some divergent trends with respect to age are observed in the NSW records (failure rate for cast iron pipes decreases with the age for a subset of pipes).

To establish the effectiveness of the proposed techniques, it is important to know how many failures can be detected by inspecting the first few pipes within a group in the ranking order. In order to establish this, we studied the prediction data for the highest ranked 1\% and 5\% of pipe lengths in each year for each state. A sample of these observations is shown in Figure \ref{AuCFig}(g)-(i). It is observed that using the RSF method has an advantage over using other methods in terms of detecting the most number of failures with a least amount of inspection effort.

We performed further analysis of predicted results using the RSF technique on the suburb level. For this study, the suburbs within a state are first sorted, based on the cumulative probability of all the pipes within the suburb for a given prediction year. This forms a suburb based dataset and the general procedure is then followed to obtain the suburb based ROC curve. Results for the VIC pipe dataset for the years 2015, 2016 and 2017 are shown in Figure \ref{SuburbModel} (a). Based on the results, it is evident that by inspecting the pipes in the top 10 suburbs, it is possible to detect more than $60\%$ of the total failures. Additionally, inspecting the pipes in only half of the total suburbs in the ranked order will result in detecting more than $80\%$ of the total pipe failures for all years.

Extending our analysis of the prediction of pipe failures in suburbs, we turn our attention towards predicting critical suburbs within each of the three states analysed. It is extremely valuable for water management authorities to be able to restrict attention in specific years to monitoring only a select number of pipe systems located in specific suburbs. This comes with the benefits of reduced manpower and labour costs. We define critical suburbs within a specific year in our model to be suburbs that host a number pipe failures greater than the average for that year. Using this definition, we use our trained models to generate with aggregate failure probabilities for each suburb. Using these aggregate probability values and the actual number of failures occurring in each suburb in that year, we calculate the precision and the recall rates for the detection of critical suburbs. As shown in Figure \ref{SuburbModel} (b), all three techniques we used as candidates demonstrated satisfactory recall (18.18\%) and precision (96.67\%) levels for suburbs in VIC for 2017. We also note that such similar behaviours were observed in the three techniques across years and geographic locations.

Our experimental results indicate that for most of the studied scenarios, the RSF technique outperforms other machine learning techniques, clearly highlighting its superior prediction capabilities in long-term predicting pipe failures. We also note that all three techniques perform better than the baseline as predicted by the historical number of previous failures in pipes. In general, the RSF technique tends to improve in its prediction accuracy or at least maintain the same accuracy as predictions are made further into the future. In contrast, we observe that the accuracy of the predictions made through GB and RF methods tend to deteriorate over time.

We finally observe that all the techniques used saturate at a maximum AUC level of around 0.65 to 0.75. While pipe failures may be extremely unpredictable, this to also due to the fact that our predictions are made further away into the future (5-7 years), and also due to the fewer number of features we use for predictions as compared to other similar studies \cite{Kumar2018}.


In the suburb based analysis, we notice that by the aggregation of pipe failure probabilities predicted by RSF across suburbs, we were able to predict a significant proportion of pipe failures. In Figure \ref{SuburbModel} (a), we clearly see that at 20\%, more than 40\% of the pipe failures were recovered. We also note that the curve for 2017 dominates that portion of the graph, again signalling the efficiency of the RSF technique for predicting failures further into the future. Our analysis of critical suburbs also reveals quite interesting facts regarding pipe failures and their distribution across suburbs. As Figure \ref{SuburbModel} (b) clearly demonstrates, all three methods perform well in precisely recalling suburbs with greater-than-average numbers of failures in a year. This coupled with the results of Figure \ref{SuburbModel} (a) also suggest that in a given year, the pipe failures are clustered in a few vulnerable suburbs within each state.


Prediction uncertainty for the long-term failure prediction also has been calculated. These long-term prediction curves show that actual failure rates align with the results generated by RSF and within the uncertainty interval. The mean failure prediction is ideal for modelling future behaviour of pipeline network benchmarking performance indices such as unplanned water interruption and water main breaks. In addition to the mean prediction, all the water utilities require uncertainty interval in order to evaluate the impact and cost of more targeted water network levels of service inform both short and long-term renewals budgets.


\section{Discussion on RSF for long-term pipe failure prediction}
The empirical study we conducted here shows that,  the longer we predict into the future, more degradation in the accuracy can be observed in RF and GB, whereas RSF remains quite consistent in the accuracy when we predict further into the future. This is because RSF explicitly takes, the time until the occurrence of a failure (Eq. \ref{SF}), into account when calculating the CHF,  making RSF to provide a robust prediction over a longer period of time. Secondly, RSF seeks a model that best explains the data and thus represents a suitable tool for exploratory analysis where prior information of the survival data is limited (consider the experimental results for QLD dataset, where we have only one year worth data for training and also the failure data is highly sparse). Thirdly, in case of multi-dimensional data, limitations of univariate regression approaches (i.e. Weibull method) such as unreliable estimation of regression coefficients or convergence problems do not apply to RSF. To the best of our knowledge, this is the first research conducted to explore the potential advantages of using RSF in pipe failure predictions along with the uncertainty estimation.

Currently, our predictive data analytic models are deployed in the city of Sydney, the region west of Melbourne and south-east QLD mainly for short-term prediction purposes. Each of these Australian water utilities are monitoring the number and the location of water main failures to validate our model. They also use our model in their internal financial modelling, risk distribution assessment planning and also to assist in the development of condition assessment programs. In addition to our previous work which have been deployed already, the study presents in this paper focuses on the development of a nonparametric survival analysis technique to determine which water main assets and suburbs are most likely to have water main failures in the next 5–7 years. Our results indicate that RSF opens up a new avenue for robust pipe failure prediction

\section{Conclusions}
The reliability of the water distribution network in any city is critical to delivering clean water supply to customers. Tailoring data science techniques to model the pipeline failure prediction provides accurate insights into water main networks. This will essentially assist water authorities to carry-out proactive pipeline maintenance. Therefore, we have presented a thorough survey of the landscape of nonparametric survival analysis as it pertains to predictions of survival rates and correspondingly decease rates of assets. We have used data from the water management authorities of three major Australian states to validate the survival analysis technique we propose, Random Survival Forest, to compare against other state-of-art machine learning techniques that have been proven effective the in similar applications. We perform a thorough analysis of the performance of the techniques in making predictions over multiple years. The results show that the Random Survival Forest (RSF) has consistently shown to outperform the other techniques, in long-term forecasting. To the best of our knowledge, this is the first research conducted to explore the potential advantages of using RSF in pipe failure predictions along with the uncertainty estimation.
Ultimately, we believe this work, at the intersection of Machine Learning and Asset Management, will lead to more effective and proactive infrastructure maintenance in the water industry across the world.

\section{Acknowledgement}
We sincerely thank Australian water utilities: Sydney Water, UnityWater and Western Water  for sharing data, expert domain knowledge and the valuable feedback.

\bibliographystyle{splncs04}
\bibliography{reference}

\end{document}